# Electric Vehicles coordination for grid balancing using multi-objective Harris Hawks Optimization

Cristina Pop[1], Tudor Cioara[1], Viorica Chifu[1], Ionut Anghel[1], Francesco Bellesini[2]

[1]Computer Science Department, Technical University of Cluj-Napoca, Memorandumului 28, 400114 Cluj-Napoca, Romania

{cristina.pop; tudor.cioara; viorica.chifu; ionut.anghel; }@cs.utcluj.ro

[2]EMOTION S.r.l, Via Gallipoli, 51, 73013 Galatina, Lecce, Italy

francesco.bellesini@emotion-team.com

**Abstract.** The rise of renewables coincides with the shift towards Electrical Vehicles (EVs) posing technical and operational challenges for the energy balance of the local grid. Nowadays, the energy grid cannot deal with a spike in EVs usage leading to a need for more coordinated and grid aware EVs charging and discharging strategies. However, coordinating power flow from multiple EVs into the grid requires sophisticated algorithms and load-balancing strategies as the complexity increases with more control variables and EVs, necessitating large optimization and decision search spaces. In this paper, we propose an EVs fleet coordination model for the day ahead aiming to ensure a reliable energy supply and maintain a stable local grid, by utilizing EVs to store surplus energy and discharge it during periods of energy deficit. The optimization problem is addressed using Harris Hawks Optimization (HHO) considering criteria related to energy grid balancing, time usage preference, and the location of EV drivers. The EVs schedules, associated with the position of individuals from the population, are adjusted through exploration and exploitation operations, and their technical and operational feasibility is ensured, while the rabbit individual is updated with a non-dominated EV schedule selected per iteration using a roulette wheel algorithm. The solution is evaluated within the framework of an e-mobility service in Terni city. The results indicate that coordinated charging and discharging of EVs not only meet balancing service requirements but also align with user preferences with minimal deviations. The assessment of the determined solutions' quality and efficacy shows promising outcomes, with convergence after 100 iterations reflected in a generational distance of 0.35 and a Pareto front error of 1.01, while the distribution of solutions exhibits strong hypervolume thus covering a significant portion of the objective space.

## 1. Introduction

Electric mobility is gaining traction due to global warming and climate change, with the sales of new battery electric vehicles (BEVs) rising in 2022 by over 15% relative to 2021 in the EU [1]. As a result, electric vehicles (EVs) have become the third most popular powertrain choice after gasoline and hybrid cars. The continuous rise in sales of electric vehicles can be primarily attributed to economic and regulatory factors [2, 3]. The price of lithium-ion batteries has decreased by 97% in the last 30 years due to research and development outcomes and increased usage of these batteries, which has surged from having around 1 MWh of cumulative installed capacity at the beginning of 1990 to reaching 550 GWh in 2022 solely for automotive applications [4]. Fossil fuels were the cheapest way to generate energy for a century, but the cost of solar energy has dropped by 89% in just a decade. Unlike fossil fuels or nuclear energy, renewable energy sources like wind and sunlight are free, so their operating costs are lower [5].

Regulations have been developed to combat climate change, simultaneously, with the efforts to achieve the transition to renewable energy. Regulation 2023/851 of the European Parliament makes mandatory the reduction with a 55% reduction of the new vehicles' emissions by 2030, and then a 100% reduction in 2035 [6]. In the United States, the federal Clean Air Act permits states such as New York and California to adopt a zero-emission vehicle standard [7]. New York, along with several other states, has joined an initiative aiming to deploy 3.3 million electric vehicles by 2025. Moreover, the Net Zero Emissions by 2050 Scenario serves as a normative pathway, outlining the global energy sector's trajectory towards achieving net-zero $CO_2$ emissions by 2050 [8, 9].

The increased share of renewables coincides with the shift towards EVs bringing technical and operational challenges on one hand, but in the long term could provide a solution to manage peak demand and grid balancing [10]. If the charging and discharging of EVs are coordinated a fleet of EVs could be jointly operated as a decentralized energy storage system used to guarantee the uninterrupted operation of a local energy system [11, 12]. As a result, the energy and mobility services can be coupled to provide additional revenue for vehicle owners or energy communities and a more reliable energy supply while avoiding investment in expensive devices [13]. Moreover, the uncoordinated charging operations tend to increase the demand at peak hours and may lead to overloads in distribution transformers and cables and do not take into consideration the actual energy context of the local grid. In addition to ensuring grid stability and managing peak loads, another significant benefit of implementing such coordination lies in the realm of cost optimization [10, 14]. Utility companies often employ dynamic pricing models that factor in the time of day and the overall demand on the grid, mirroring the principles of a free market where prices are determined based on the intersection of supply and demand [15]. By strategically coordinating EVs charging activities during periods of low-cost electricity, both EV owners can take advantage of reduced electricity rates, potentially leading to overall cost savings in the charging process [2, 16].

Therefore, there has been a growing interest by the research community to develop strategies to coordinate the charging and discharging of these vehicles in a way that can help to balance the supply and demand of electricity on the grid [17-19]. The latest research focuses on utilizing artificial intelligence to manage EVs for various purposes, such as integrating renewable energy, demand response, or decentralized energy storage [20]. For example, off-peak demand charging schedule the EVs for time slots when the grid load is minimal, or on renewable generation peak charging schedule the EVs for time slots when there is a high availability of renewable in the local grid. However, these techniques only partially address the issue and require interoperability with energy system operators to get the load and time of day information [21]. Vehicle-to-Grid (V2G) technology addresses the issues by enabling bi-directional communication between the EVs and the grid enabling them to feed energy back into the grid when needed by discharging the batteries [22]. This enables the smart coordination of a fleet of EVs to offer features like energy storage [23]. However, coordinating the power flow from multiple EVs into the grid requires more sophisticated grid management algorithms dealing with many variables and integrating load-balancing strategies. Smart control of the decentralized EV charging stations (CSs) can be addressed using reinforcement learning techniques that do not need explicit models e.g., typical arrival and departure times will be demonstrated [24]. However, limitations in optimization problem and decision search space sizes with the number of vehicles and considered variables is a challenge [25]. As the coordination problem involves numerous variables and constraints, metaheuristic algorithms are a suitable option due to their ability to quickly and efficiently search through a vast solution space to find optimal solutions [26].

In this paper, we address the identified challenges in the field by proposing an EVs coordination model to provide balancing services in the local grid. The model determines the optimal planning of EVs' charging and discharging for the day ahead to ensure a reliable energy supply and maintain a stable local grid, by storing the energy surplus in EVs and discharging it in periods of energy deficit. The day-ahead coordination of EVs is a multi-criteria optimization problem where each time instance represents an objective, aiming to align energy demand and supply as closely as possible considering the local operational constraints of EVs and CSs, user time of use constraints, and balancing services constraints. We have addressed it using the Harris Hawks Optimization (HHO) algorithm that mimics the hunting tactics of Harris hawks which work together in a group to catch their prey (i.e., a rabbit) by communicating and maintaining relative positions or adjusting the hunting strategy in real-time [27]. We map the Harris hawks and the rabbit to EVs schedules represented as energy matrices indicating charging and discharging activities at specific stations during different periods. In the algorithm's iterations, the EVs schedules associated with the position of individuals from the population are adjusted through exploration and exploitation operations and are further refined to ensure their technical and operational feasibility. The rabbit individual is updated with a non-dominated EV schedule selected using a roulette wheel algorithm per iteration. The collaboration mechanism allows solutions to be shared among individuals, enhancing the algorithm's ability to find optimal or near-optimal solutions in parallel.

The novel contributions of the paper are:

- A coordination model for a fleet of EVs to provide balancing services in the local grid considering operational, technical, energy service, and time of use constraints.
- A multi-objective HHO algorithm for optimal scheduling of charging and discharging actions mimics day-ahead using energy matrices and operators for efficient exploration and exploitation of the search spaces.
- Evaluation in the context of an electric mobility service from Terni, Italy, considering API interoperability for gathering the operational constraints and technical characteristics of the CSs and a fleet of electric vehicles. Therefore, the optimization model can make informed decisions based on up-to-date information, resulting in a more accurate and effective EV charging and discharging schedule to balance service requirements.

The rest of the paper is structured as follows: Section 2 presents the state-of-the-art EVs coordination models focused on the ones using nature-inspired heuristics, Section 3 describes the proposed EVs coronation model and the constraints considered, Section 4 shows how HHO was used to solve optimization problem associated with EVs coordination to provide balancing services, Section 5 describes the evaluation results in the context of a mobility service from Terni, Italy, Section 6 discusses the impact of various parameters tunning of HHO performance and converge, while Section 7 presents conclusions and future work.

## 2. Related work

The state-of-the-art literature uses artificial intelligence solutions for coordinating the charging and discharging of EVs to meet different objectives such as grid balancing, renewable energy integration, demand response, decentralized energy storage, etc. The EVs coordination problem deals with many variables and constraints therefore metaheuristics such as the genetic and Particle Swarm Optimization (PSO) are mainly used to address these issues.

Genetic optimization is used for tasks such as EVs route optimization, CS placement, and energy management to optimize energy use and minimize costs. Song et al. [28] use an evolutionary approach based on the constraint non-dominated sorting genetic algorithm (NSGA-II) to schedule the charging and discharging of EVs to maximize the benefits of all stakeholders involved, including EVs' owners, CS operators, and the grid operator. Three objective functions are considered: electricity cost minimization for owners, revenue maximization for CS operators, and energy balancing for grid operators. Morais et al. [29] combine NSGA-II with a deterministic technique to minimize both the charge cost and greenhouse gas emissions. The NSGA-II algorithm is used to generate a set of Pareto-optimal solutions, and fuzzy set theory is used to select the best solution from the Pareto set. For the considered optimization problem several constraints are defined including the power flow in the lines that must be as low as possible. Jin et al. [30] propose a power management solution to reduce the operating cost of PV-based CSs by using EVs as a decentralized energy storage system. The coordination is modeled as a multi-objective optimization considering variables such as the remaining parking time, the current and target state of charge (SoC) of the EV's battery, and the maximum number of allowable battery charging and discharging cycles. Similarly, in Poniris et al. [31] the genetic algorithm is used to minimize the overall costs associated with EVs charging and to ensure that the energy demand from the EVs does not exceed the capacity of the electrical grid. Piamvilai et al. [32] apply genetic algorithms for EV charge scheduling by considering various constraints such as bus voltages, bus powers, and line flow restrictions. The aim is to optimize the system load factor and assure user satisfaction. The scheduling algorithm takes as inputs a set of charging event matrices that are continuously updated during the charging process. The matrices store information about EVs (i.e., EV charging location, charging duration, parking duration, charging power, and initial State of Charge - SoC) and are used to optimally determine the available charging slots. Milas et al. [33] also propose a two-layer genetic algorithm to schedule the charging of electric vehicles and to generate personalized charging profiles, depending on the availability of the charger at a CS, the preferences of the driver and the CS operator. By decoupling the optimization criteria from the search space, it becomes possible to customize the criteria to meet the specific needs of each application. Wang et.al. [34] use NSGA-II multi-objective optimization to optimally schedule the EV charging to minimize the peak load and maximize the CSs benefits. Since the optimization criteria are contradictory, the purpose of the NSGA-II algorithm is to identify a compromise solution (i.e., the optimal Pareto solution) that is good enough from the perspective of both criteria. In Abdullah-Al-Nahid et al. [35] the genetic algorithm can help manage grid network stress and improve user satisfaction by scheduling EVs charging while using efficiently microgrid's renewable energy generation. The algorithm determines the best combination of EVs based on energy demand, power ratings, kWh values, and SoC values. The chosen EVs are then scheduled such that the surplus energy within the microgrid is used to the maximum using the valley filling technique.

PSO or population-based heuristics are used for similar purposes as the genetic solutions, but in their case, the optimal solution is found by iterating population individuals in the solution space based on their movements and the movements of their neighbors in search of a goal. Yang et al. [36] introduce an optimization model based on an enhanced PSO to optimize the charging and discharging behaviors of EVs in a power grid, considering several constraints: network constraints, on-load switch transformer constraints, transmission power constraints, charging and discharging power constraints, and SoC, etc. The optimization objectives are reducing power losses, fine-tuning the frequency of control equipment within the power network, achieving a smoother daily load curve, and ensuring the satisfaction of the electric vehicle owner. A feasible reservation strategy is applied to handle the inequality constraints related to charging and discharging and the number of particles in the swarm. Konstantinidis et al. [37] apply PSO to schedule the charging of EVs within a parking lot and minimize the overall charging cost while respecting technical constraints referring to the frequency of EV charging and discharging, boundaries for battery energy levels and charging power, the EV's arrival and departure time, and the number of EVs engaged in the scheduling. Similarly, in Fernandez et al. [38], PSO and mixed frog jumping algorithms are applied to minimize the total cost of grid charging, considering demand dynamics, electric vehicle arrival and departure times, and prioritizing EVs that have booked a charging slot. Mohammad et al. [39] use a binary particle swarm optimization algorithm for optimal scheduling of charging and discharging EVs as a distributed energy storage within the grid. The optimization criteria include minimizing energy costs, managing peak load demands, extending the lifespan of batteries, and fulfilling the travel needs of users. Fang et al. [40] solve the problem of optimal energy scheduling of EVs connected to a distribution network using an improved particle swarm optimization algorithm. The problem is modeled as a multi-objective optimization to minimize the associated cost with EVs, and the cost attributed to battery degradation. The PSO is used to reduce root-mean-square erosion as well as peak-valley differentiation of the system. In Wang et al. [41], a multi-objective particle swarm optimization is applied for the energy scheduling of EVs within urban residential zones to reduce the total power load and scheduling cost. The solution advantage is the few adjustable parameters, fast convergence speed, and strong optimization capability while considering factors related to the operation mode of the PV system, EV charging requirements, transformer capacity limitation, SoC, and user behavior consideration. Savari et al. [42] use Particle Swarm Optimisation for optimal charging of EVs as well as other solutions such as Arrival Time-Based Priority and SOC-Based Priority. Their objective is to minimize the cost of charging while meeting constraints related to the state of the CS while comparing the performance of the solutions proposed. They are considering microgrids with the renewable resources scenario to reduce energy consumption and charging costs by strategically redistributing EV load and to enhance the customer's benefits by reducing the electricity cost.

Other heuristic algorithms applied for optimal energy scheduling of EVs are Improved Marine Predator, Artificial Bee Colony, hybrid metaheuristics, greedy heuristic, etc. Sowmya et al. [43] use the Improved Marine Predator algorithm to reduce the cost of operating EVs in the microgrid. The algorithm considers the uncertainties due to changes in battery performance and electricity pricing by iteratively refining the scheduling plan using an opposition-based learning schema. The constraints deal with the SoC of the batteries and the reduction of the frequency of charging/discharging cycles, etc. Álvarez et al. [44] apply a hybrid artificial bee colony algorithm to meet the energy demands and maximize the available power usage while managing a fleet of EVs. The algorithm considers two cases: static scheduling with a priori known energy requirements of EVS and dynamic scheduling, with energy requirements adaptable to changing energy demands. The proposed algorithm outperforms other state-of-the-art techniques on the benchmark proposed in Hernandez-Arauzo et al. [45]. Ahmadi et al. [46] address the problem of optimal allocation of

EVs parking lots and the optimal scheduling of electric vehicles in a smart distribution network as an optimization problem solved using a hybrid metaheuristic algorithm that integrates the genetic, PSO, and Imperialist Competitive Algorithms. The genetic algorithm is initially used to identify the optimal solution for the grid losses and voltage drop which is then improved by applying PSO and Imperialist Competitive Algorithms. The proposed solution aids utilities and transportation agencies in improving EV operation and reducing charging's impact on the power grid. Abdel-Hakim et al. [47] use a greedy heuristic to optimally schedule the EVs while considering the uncertainty in energy generation. The objectives are to minimize the overall energy cost and balance the microgrid electricity by considering electricity price, EVs' arrival and departure times, and total revenues to meet load requirements. Limmer et al. [48] proposed a large neighborhood search algorithm to schedule the charging of a fleet of EVs, considering reservations made and energy available on the grid. The algorithm can iteratively destroy and restore the components of the best solution identified so far by introducing three destroy operators (i.e., random, relatedness, and no overlap operators) and a repair operator using mixed integer linear programming.

## 3. EVs coordination model

We aim to optimize EVs' charging and discharging schedules to balance the local power grid. This will allow a fleet of EVs to provide balancing services and accommodate excess renewable energy while maintaining safe operations and energy supply.

### 3.1. Assumptions and constraints

We denote with $CS$ the set of EV charging stations connected to the grid:

$$CS = \{cs_j | j \in \{1 \dots m\}\} \quad (1)$$

where $m$ represents the number of functional CSs. A charging station $cs_j$ is characterized by the following features represented as variables of a decision vector for each time slot $t$:

$$cs_j = [P_j, type_j, L_j]^t \quad (2)$$

where $P_j$ is the charging or discharging power of the V2G station, and $L_j$ is the station location of the station in the local grid. For type of the station ($type_j$) parameter we use Level 3 stations that are DC- very fast CSs that can charge batteries to 80% in 20-40 minutes.

The charging or discharging power of the CS is an important feature of it since different models of electric cars have different charging power constraints. It is determined based on the nominal voltage ($V_j$) and intensity of the current ($I_j$) characteristics of the station:

$$P_j = V_j * I_j \quad (3)$$

Consequently, when an EV needs to be scheduled for charging/discharging, the station is chosen considering the maximum AC or DC power available for the slot and time needed for charging or discharging.

Other factors that may be considered are the weather conditions, planed time slots for the next day, EV's battery lifespan (i.e., number of charging/discharging cycles remaining), and the charging level of the battery of the EV. If the state of charge (SoC) of the battery is below 20% or above 80%, the charging speed is lower, to increase the battery life.

The fleet of EVs is denoted with:

$$EV = \{ev_i \,|\, i \in \{1 \dots n\}\} \quad (4)$$

where $n$ is the number of EVs in the fleet and $ev$ is formally defined based on its changing features at time slot $t$ as:

$$ev_i = [SoC_i, E_{i,c}, E_{i,d}, lifespan_i, t_i, \phi_i]^t \quad (5)$$

where: $E_{i,c}$ and $E_{i,d}$ is the amount of energy that can be charged or discharged for the battery, $t_i$ is the time needed to operate charging or discharging and $\phi_i$ is the location in the microgrid and $lifespan_i$ is the battery lifespan. We have considered that the EVs maximum charging or discharging time $t_i$ is always lower or equal to the time slot $t$ used in scheduling:

$$t_{i,max} \leq t, (\forall)t \in T, t = hour, T = day \tag{6}$$

The *state of charge* $SoC_i$ of $ev_i$ battery are expressed in percentages relative to EV maximum capacity $C_{max}$ specified by manufacturers. Thus, $SoC$ value of 100% means that the battery is fully charged, while $SoC$ value of 0% means that the battery is completely discharged. As the battery's maximum capacity $C_{max}$ decreases over time, a $SoC$ value of 100% on an older battery may only represent a fraction of the capacity compared to a new battery.

Considering the relation (3) the amount of energy that can charged from the CS in the time slot $t$ is calculated as:

$$cs_{j,E}(T) = U_j * I_j * t \tag{7}$$

The energy to be charged in the he $ev_i$ EV's battery is determined as [49]:

$$E_{i,c}(t_i) = (SoC_{i,G} - SoC_{i,c}) * \frac{I_{max}}{100} * t_i \tag{8}$$

where $SoC_{i,G}$ is the target value of the battery state of charge, $SoC_{i,c}$ is the current value of the state of charge, and $I_{max}$ is the maximum intensity of the current needed to reach the maximum capacity of the battery:

$$I_{max} = \frac{C_{max}}{U_j * 1000} \tag{9}$$

The energy discharged from an EV's battery can be estimated similarly also adding the loss ($\sigma$) of energy of the conversion process for V2G technology:

$$E_{i,d}(t_i) = (1 - \sigma) * E_{i,c}(t_i) \tag{10}$$

The lifespan of the battery may influence its scheduling decision for charging and discharging. We have considered that the number of remaining charge-discharge cycles of the battery, $n_{i,r}$, takes values in the following interval:

$$0 \leq n_{i,r} \leq n_{i,max} - \tau \tag{11}$$

where $n_{i,max}$ is the maximum number of charge-discharge cycles of the EV's battery specified by the manufacturer for correct exploitation, and $\tau$ is an integer value used to approximate the impact of partial charging and discharging operations on the battery's remaining useful life. The remaining charge-discharge cycles of the battery can be approximated as:

$$n_{i,r} = n_{i,max} - n_{i,c} \tag{12}$$

where $n_{i,c}$ is the number of charge-discharge cycles performed up to the current time.

The time to charge or discharge an EV's battery from the actual SoC to the target one is bounded by a minimum and maximum operational time:

$$t_{i,min} \leq t_i \leq t_{i,max} \tag{13}$$

The minimum charge or discharge time, $t_{i,min}$ reflects the time needed to reach the $SoC$ target value under normal conditions and the maximum time $t_{i,max}$ reflects the impact of special not optimal conditions:

$$t_{i,min} = \frac{C_{max}}{P_j * PF} * \frac{SoC_{i,t} - SoC_{i,c}}{100} \tag{14}$$

$$t_{i,max} = \frac{C_{max}}{P_j * PF} * \frac{SoC_{i,t} - SoC_{i,c}}{100} + \Delta t \tag{15}$$

where $P_j$ is the charging power of the station, $PF$ is a factor that specifies the energy efficiency of the power transfer from the charger to the EV, and $\Delta t$ is a value that reflects an increase in the operation time due to special conditions such as extreme weather conditions, battery level, etc.

The estimated waiting time for an EV, $ev_i$, before charging or discharging at station $cs_j$ is defined as:

$$t_{wait}(ev_i, cs_j) = \sum_{k=1}^{K} t_k \tag{16}$$

where $t_k$ is the charging or discharging time of EVs scheduled before $ev_i$ at $cs_j$:

$$scheduled(ev_k, cs_j) > 0 \ \& \ t_k < t_j, \ (\forall) \ t_k, t_j \in T \tag{17}$$

Finally, we have considered the constraints of the citizens driving the cars related to the $ev_i$ scheduling at CSs, $\varphi_i$ such as the time interval, waiting time, distance to the station location, etc. The preferences are ordered based on their importance (i.e., low, medium, high) using a relation $\succ$ defined on a set $\varphi_i$ as:

$$\varphi_1 \succ \varphi_2 \ if \ \varphi_1 \succcurlyeq \varphi_2 \ \& \ \varphi_2 \not\succcurlyeq \varphi_1 \tag{18}$$

meaning that preference $\varphi_1$ should be considered before $\varphi_2$. The relation over the preferences set defined in our model satisfies the completeness and transitivity axioms:

$$(\forall) \ \varphi_1, \varphi_2 \in \phi_i, \ \varphi_1 \succcurlyeq \varphi_2 \otimes \varphi_2 \succcurlyeq \varphi_1 \tag{19}$$

$$(\forall) \ \varphi_1, \varphi_2, \varphi_3 \in \phi_i, \ if \ \varphi_1 \succcurlyeq \varphi_2 \wedge \varphi_2 \succcurlyeq \varphi_1 \ then \ \varphi_1 \succcurlyeq \varphi_3 \tag{20}$$

For example, preference for the time slot $t_i$ during the next day, $\theta$ is expressed as:

$$\varphi = t_i: t_i \succ t_j, (\forall) \ t_j \in \theta \wedge t_i \in \theta \wedge priority(t_j) = high \tag{21}$$

### 3.2. Energy scheduling optimization

In light of the above-defined model assumption and constraints, the scheduling of EVs charging and discharging actions at stations be optimally coordinated to deliver balancing services energy services in a renewable-powered local grid. We consider local grids featuring small-scale renewable generation resources ($Res$), consumers ($C$), and EVs that are used as second-life energy storage devices:

$$SG = (Res, C, EVs) \tag{22}$$

The balancing services are defined for the next day by looking at the forecasted total energy demand and generation values for the microgrid over the optimization window $T$. The difference between the two energy profiles over $T$, gives the energy balance profile of the local grid considering also the uncertainty associated with the estimates $U^*_{prediction}$ over interval $T$:

$$E_{balance}(T) = E_{res}^{prediction}(T) - E_c^{prediction}(T) + U^*_{prediction}(T) \ \forall t \epsilon T \tag{23}$$

To model the uncertainty of the energy prediction we have used the value-at-risk method. It measures with a confidence level the maximum loss of a of the EVs scheduling optimization due to inaccurate estimation of the energy balance. Historical values of energy balances can be used to determine the arithmetic return over the period $[i, i + 1]$:

$$R_i = \frac{E_{balance}^{i+1}(T) - E_{balance}^{i}(T)}{E_{balance}^{i}(T)}, 0 \leq i \leq d \tag{24}$$

where $d$ is the number of past days considered. On the sorted sets of returns $\check{R}$ the value $\check{R}_\alpha$ corresponding to the confidence level α is determined as:

$$VaR = \mu(\check{R}) - \check{R}_\alpha \tag{25}$$

where $\mu(\check{R})$ is the mean return of the determined set, and $\check{R}_\alpha$ represents the worst return from the set. The predicted energy balance can have either a positive or negative value depending on the existence of a surplus or deficit of energy at the grid level at time slot $t$:

$$\begin{cases} e_{balance}(t) > \varepsilon, coordinated\ EVs\ charge \\ e_{balance}(t) \cong 0,\ no\ actions \\ e_{balance}(t) < \varepsilon,\ coordinated\ EVs\ charge \end{cases} \tag{26}$$

To maintain a balance between the generated energy and the produced energy at the level of the grid over next-day interval $T$ we need to schedule a set of EVs for charging or discharging at the set of stations, such that the distributed energy storage profile $E_{storage}(T)$ compensate the surplus or deficit predicted:

$$E_{storage}(T) = \{e_{storage}(t) | t \in T\} \tag{27}$$

In each moment the amount of distributed storage of energy is determined based on the charging and discharging actions on available EVs:

$$e_{storage}(t) = \sum_i (E_{charge}(ev_i, t) - E_{discharge}(ev_i, t)) \tag{28}$$

The operations of charging or discharging the battery for an EV, $ev_i$, are mutually exclusive for a time slot $t$ meaning that $ev_i$ is either charged or discharged.

In our scheduling problem, this is translated into an objective function associated with the energy service to be delivered by the fleet of EVs:

$$e_{balance}(T) + e_{storage}(T) \cong 0, (\forall)\ t \in T \tag{29}$$

For the next day, $T$ EVs should be scheduled to either charge or discharge to meet the objective function defined in relation (29). We have addressed this by defining an EVs scheduling matrix $s$ having columns equal to the time slots $t$ of the next days while the rows are equal to the $cs_j$ available in the grid (see Table 1).

Table 1: EVs scheduling matrix representation.

|    |        | $t_1$ | $t_2$ | ... | $t_p$ |
|----|--------|-------|-------|-----|-------|
|    | $cs_1$ |       |       |     |       |
| CS | $cs_2$ |       | $< ev_i, E_{i,c}, E_{i,d} >$ |     |       |
|    | ...    |       |       |     |       |
|    | $cs_m$ |       |       |     |       |

Table header spans: T covers $t_1$, $t_2$, ..., $t_p$.

As the charging and discharging operations are mutually exclusive in each cell of the matrix, we will have either a positive energy value corresponding to specific EV scheduling on the time slot:

$$s(CS,T) = \begin{bmatrix} s_{1,1} & s_{1,2} & \dots & s_{1,t} \\ \dots & \dots & \dots & \dots \\ s_{j,1} & s_{j,2} & \dots & s_{j,t} \end{bmatrix}, (\forall)\, j \in \{1,..m\}, (\forall)\, t \in T \qquad (30)$$

$$s(j,p) = \pm E_i, E_i = either\ E_{i,c}\ or\ E_{i,d}\ for\ ev_i, (\forall)\, i \in \{1,..n\} \qquad (31)$$

Summing the elements of the matrix $s$ on each column we will get the value of corresponding $e_{storage}(t)$:

$$e_{storage}(t) = s(CS,t) = \sum_{j=1}^{m} s(j,t) \qquad (32)$$

On the EV scheduling matrix, several constraints need to be enforced. The energy storage amount cumulated for each time slot $t$ must be as closed as possible or higher than the energy balance:

$$0 \le e_{storage}(t) > e_{balance}(t) \qquad (33)$$

Also, there cannot be two EVs scheduled at the same CS in the same time slot t:

$$(\forall)\, cs_j, t, if\ s(cs_j, t) > 0\ and\ ev_i, ev_z\ scheduled\ at\ cs_j\ at\ t\ then\ i = z \qquad (34)$$

A car cannot be scheduled for discharge if the maximum number of remaining charge-discharge cycles of the battery is less than a predefined value, ε specified by the manufacturers.

$$(\forall)ev_i \in EV, n_{i,r} \ge \varepsilon \qquad (35)$$

Finally, the distance between the CS location and the EV's location should be below a predefined threshold $l$:

$$(\forall)\, ev_i, cs_j, t, s(cs_j, t) = ev_i \Leftrightarrow |cs_{j,L} - ev_{i,L}| \le l \qquad (36)$$

## 4. Multi-objective HHO for grid balancing

The population of Harris hawks are modelled as agents each featuring a position in the search space $X(i)$ which may change in each iteration $i$. We have adapted the optimization process of Heidari et al. [27] that includes phases of exploration, transition, and exploitation, and relevant operators to address the specificity of EVs scheduling and coordination to provide balancing services.

In our case, we have modeled the position of the Harris hawks using the scheduling matrix of EVs on the CSs:

$$X(i) = s(CS, T) \qquad (37)$$

As the objective is to find the best scheduling of EVs (i.e., the $s$ matrix) to balance the grid energy generation and demand the fitness function will be based on the relation (29):

$$f_t = \min_{s(CS,t)} \left( e_{balance}(t) + e_{storage}(t) \right) \forall\, t \in T \qquad (38)$$

We model the prey as the agent with the best position in the iteration $i$ determined using our fitness functions over the entire day ahead optimization window:

$$X_{prey}(i) = \min f_T(X^N(i)) \, \forall\, t \in T \qquad (39)$$

where $N$ is the number of agents of the population.

In the exploration phase, the algorithm searches for new solutions by improving the best one found so far. Each solution represents a Harris hawk location, and the best one is seen as the prey. The search is guided by relations that describe how hawks detect and locate prey:

$$X(i+1) = \begin{cases} X_{rand}(i) - r_1 * |X_{rand}(i) - 2 * r_2 * X(i)| & q \geq 0.5 \\ \left(X_{prey}(i) - X_m(i)\right) - r_3 * \left(LB + r_4 * (UB - LB)\right) & q < 0.5 \end{cases} \quad (40)$$

In relation (40), $X(i+1)$ is the updated position of the hawk in the next iteration $(i+1)$, $X(i)$ is the current position of the hawks, $X_{rand}(i)$ is a hawk selected randomly from the current population, $X_{prey}(i)$ is the position of the prey, $r_1, r_2, r_3, r_4 \in (0,1)$ are random numbers, $UB$ and $LB$ are the upper and lower bounds of variables. The first rule ensures that the algorithm can find a good solution even if it starts from bad positions while the second rule adjusts the gap between the position of the current best solution and the average position of the rest of the individuals. The selection of rules is driven by $q$ a random number in the interval $(0,1)$.

The average of the positions of all the hawks, $X_m(t)$ is, computed as:

$$X_m(t) = \underset{for \neq 0 \text{ cels in } s}{\text{avg}} |X_n(i)|, n = 1 \ldots N \quad (41)$$

where $N$ is the total number of hawks of the current population in iteration $i$.

In the transition phase, the algorithm models the escape energy of the prey, which influences the behavior of the hawks to go from the exploration phase to the exploitation phase:

$$E = 2 * E_0 \left(1 - \frac{i}{\theta}\right) \quad (42)$$

where $E$ is the prey energy, $E_0$ is the initial energy of the prey in each iteration $i$ that is randomly generated between -1 and 1 and $\theta$ is the maximum number of iterations. Prey strength is based on energy compared to previous generations. The increase makes them stronger while the decrease makes them weaker. If the absolute value of $E$ is greater than 1, the exploration phase continues, and if the absolute value of $E$ is less than or equal to 1, the exploitation phase begins.

The *exploitation phase* models the hawks behavior as they pounce on the prey they have explored. In this phase, four different chasing strategies can be used for the selection depending on values of escaping energy, $E$ and the chance of escaping, $r$.

The *soft besiege strategy* occurs when the prey has some energy and tries to escape the hawks. In response, the Harris hawks will surround the prey to tire it out, and then launch a surprise attack. This strategy is used when the chance of successful escape, $r \geq 0.5$, and the escaping energy of the prey $E \geq 0.5$. The process of the soft besiege strategy is modeled as:

$$X(i+1) = (X_{prey}(i) - X(i)) - E * |J * X_{prey}(i) - X(i)| \quad (44)$$

$$J = 2 * (1 - r_5) \quad (45)$$

where $J$ represents the prey escaping $r_5 \in (0,1)$ is a random number.

The *hard besiege strategy* occurs when the prey is exhausted and unable to escape, and the hawks capture it using a surprise attack. The process of the hard besiege strategy is modeled as:

$$X(i+1) = X_{prey}(i) - E * |X_{prey}(i) - X(i)| \quad (46)$$

The *soft besiege strategy with progressive rapid dives* occurs when the prey has enough energy to escape, $|E| \geq 0.5$, and the chance of escape is $r < 0.5$. In this case, the hawks will surround the prey less aggressively, trying to gradually drain the prey's energy and thus increasing their chances of capture. The Harris hawks use a feedback mechanism to adjust their movement by comparing the fitness of their current position with the previous one. If the current fitness $f$ is not improving a random walk is done to improve their chances of capturing the prey ($M$):

$$X(t+1) = \begin{cases} X_{prey}(t) - E * |J * X_{prey}(t) - X(t)|, & if\ f < f(X(t)) \\ X_{prey}(t) - E * |J * X_{prey}(t) - X(t)| + M, & if\ f < f(X(t)) \end{cases} \quad (47)$$

The *hard besiege with progressive rapid dives* occurs when the energy of the prey, $|E| < 0.5$, and the chance of escape, $r < 0.5$. The hawks focus on reducing the distance between their average position, $X_m$, and the prey position $X_{prey}$:

$$X(t+1) = \begin{cases} X_{prey}(t) - E * |J * X_{prey}(t) - X_m(t)|, & if\ f < f(X(t)) \\ X_{prey}(t) - E * |J * X_{prey}(t) - X(t)| + M, & if\ f < f(X(t)) \end{cases} \quad (48)$$

The strategy is more aggressive and aims to quickly reduce the prey's energy and increase the chances of a successful capture.

The fitness function (38) involves minimizing the distance between two energy profiles $e_{balance}$ and $e_{storage}$ over an optimization interval $T$, thus optimization should be done for each time slot $t$. As we are dealing with a multi-criteria optimization problem, we aim to find Pareto optimal solutions that are good enough for the majority of time slots $t$ of $T$:

$$f_T(X(i)) \xrightarrow{yields} < f_{t_1}(X(i)), f_{t_2}(X(i)) \dots f_{t_p}(X(i)) >, \ t_1, t_2, \dots t_p \in T \quad (49)$$

In this scope, we define a mechanism for achieving the non-dominant solutions identified using a dominance relation in each iteration. A solution $X(i)$ dominates another solution $X'(i)$ if, for all sub-objectives $f_{t_p}$, the value of $X(i)$ is not worse than the value of $X'(i)$, and for at least one objective, the value of $X(i)$ is strictly better:

$$(\forall) p \in T, f_{t_p}(X(t)) \leq f_{t_p}(X'(t)) \wedge (\exists) p\ f_{t_p}(X(t)) < f_{t_p}(X'(t)) \quad (50)$$

In each iteration, the archive is updated by comparing new solutions found with the solutions already saved in previous iterations. If a new solution is dominated by an existing solution, it is not added to the archive. If it dominates one or more existing solutions, the dominated solutions are removed, and the new solution is added to the archive. If neither the new solution nor existing solutions dominate each other, the new solution is added to the archive.

In case the archive is full, the solutions that need to be replaced are randomly selected from the subset of solutions having low crowding distance values (see Algorithm 1).

**Algorithm 1: Determine crowding distance value**
**Inputs:** $A -$ archive of non-dominant solutions, $N$ - the number of solutions in the archive
**Outputs:** $M = Map(D_c[i], X(i))$ – the map of distances for all solutions in the archive
**Begin**
1       **Foreach** solution $X(i)$ in archive $A$ **do**
2          $D_c[i] = 0$
3       **Foreach** objective function $f_{t_p} \in f_T$ **do**
4          **Foreach** solution $X(i)$ in archive $A$ **do**
5            $S_A = Sort_{descending}(X(i), f_{t_p}))$
6       **End foreach**
7       **End foreach**
8       $D_c[0] = \infty\ M.put(D_c[0], S_A[0])$
9       $D_c[n] = \infty\ M.put(D_c[n], S_A[n])$
10      **For** $j = 1$ to $(n-1)$ **do**
11         $S_A[j] = S_A[j] + \frac{f_{t_p}(S_A[j+1]) - f_{t_p}(S_A[j-1])}{f_{t_p}^{max}(S_A(j)) - f_{t_p}^{min}(S_A(j))}$
12         $M.put(D_c[j], S_A[j])$
13      **End for**
**End**

The crowding distance provides an estimate of the density of the solutions in the neighborhood of a solution in the objective space [24]. A high crowding distance value indicates that there are fewer solutions in the vicinity of that solution, while a low crowding distance value indicates that there are many solutions in the vicinity of that solution. The solutions, which have the smallest and largest objective values, are assigned an infinite crowding distance to ensure that they are always selected for the next generation (see lines 3-9). The distance for the intermediary solution is determined on the algorithm line 11.

We use the roulette wheel method [50] to select the prey position ensuring that the prey is chosen from less populated areas of the archive. The solution with the highest crowding distance value is more likely to be selected as the prey thus increasing the diversity of the population and avoiding premature convergence to suboptimal solutions.

The multi-criteria HHO solution for optimal scheduling the charging and discharging of EVs at the CSs to deliver balancing services to the grid is presented in Algorithm 2. It starts by generating the initial population of candidate solutions and the initial positions of individuals (see lines 1-4), iteratively updating the population of candidate solutions in each iteration until the maximum number of iterations is reached (lines 5-34) using the defined relations and the techniques for constraints, preferences, and solution dominance handling, select the best solution from the final population of candidate solutions as the solution to the optimization problem (line 35).

**Algorithm 2: Multi-criteria HHO for EVs coordination**

**Inputs:** $N$ - Population size, $max_i$ - maximum number of iterations, $i$ – the iteration number, $E_0$ – the initial energy of the prey, $A$ – the archive of non-dominant solutions and $n$ the maximum dimension, $EV$ - the set of electric vehicles, $CS$ – the set of CSs, $e_{balance}(T)$ – the grid energy balance profile over the optimization window $T$

**Outputs:** $X_{sol}(i)$, the solution corresponding to the EVs scheduling matrix s with the best fitness

**Begin**
1   Generate the initial population $P = \{X^0(i), i = 1,2 \dots, N, |X^0(i) = s(EV, CS)\}$
2   **Foreach** $X^0(i)$ **do** $F_T(X^0(i))$ **End foreach**
3   $A = nonDominated\left(X^0(i), f_{t_p}\right), f_{t_p} \in F_T$
4   $i = 0$
5   **while** ($i < max_{iter}$)
6      **Foreach** $X(i) \in A$ **do**
7         Compute the crowding distance $D_c$ values using Algorithm 1
8         $Sort_{descending}\ (X(i), D_c)$
9      **End foreach**
10   $X_{prey}(i) = RouletteWheel\ (A, [D_c[n]])$
11   **Foreach** solution $X(i)$ in $P$ **do**
12      Update the energy value $E$ for $X(i)$ using relation (47)
13      **if** $|E| \geq 1$ **then** update $X(i)$ using relation (44)
14      **elseif** $|E| < 1$ **then**
15         if $r \geq 0.5\ \&\ |E| \geq 0.5$ then update $X(i)$ using relation (48)
16         if $r \geq 0.5\ \&\ |E| < 0.5$ then update $X(i)$ using relation (50)
17         if $r < 0.5\ \&\ |E| \geq 0.5$ then update $X(i)$ using relation (51)
18         if $r < 0.5\ \&\ |E| < 0.5$ then update $X(i)$ using relation (52)
19      **End if**
20      Compute the fitness values $F_T(X(i))$
21   end foreach
22   **If** $nonDominated\left(X(i), f_{t_p}\right), f_{t_p} \in F_T$ **then**
23      **If** $!\ isFull\ (A)$ **then** insert in $X(i)$ in $A$
24      **Else**
25         foreach $X(j) \in A$ **do**
26            Compute the crowding distance $D_c$ values using Algorithm 1
27            $Sort_{descending}\ (X(j), D_c)$
28            Randomly select $X(j)$ having low distance and replace it with $X(i)$ in $A$
29         **End foreach**

| 30 | End if |
| 31 | End if |
| 32 | $i = i + 1$ |
| 33 | $X_{prey}(i) = RouletteWheel\ (A, [D_c[n]])$ |
| 34 | **end while** |
| 35 | **return** $X_{sol}(i) = best\ F_T(X(i)),\ X(i) \in A$ |
| **End** | |

## 5. Evaluation results

To validate the proposed solution, we have leveraged the data from Terni city, Italy, provided by Emotion, a CSs operator and e-mobility service provider from Italy [51]. They offered access to 10 physical EVs (see Tables 2 and 3) equipped with an on-board diagnostic device, providing the following data every 5 seconds using the MQTT protocol: vehicle ID, brand and model, current battery capacity and SoC, velocity, total kilometers displayed by the odometer, amount of last charged energy, the discharging speed, and the timestamp when the data has been collected. Table 2 shows a sample of the EVs real-time data collected.

*Table 2: Examples of data from EVs collected using API.*

| Vehicle ID | Model | Charging Power (kW) | Battery Capacity (kWh) | Measure ID | Autonomy (Km) | SoC (%) | Position Latitude | longitude | Timestamp |
|---|---|---|---|---|---|---|---|---|---|
| FE132DG | Renault Zoe | 22 | 41 | 156294 | 166 | 57 | 43.1445 | 12.4496 | 2/20/2020 7:10 |
| | | | | 156297 | 155 | 55 | 43.1445 | 12.4495 | 2/20/2020 8:15 |
| | | | | 156300 | 144 | 52 | 43.0721 | 12.3421 | 2/20/2020 9:06 |
| | | | | 156302 | 156 | 56 | 43.0721 | 12.3420 | 2/20/2020 9:48 |
| | | | | 156304 | 207 | 74 | 43.0782 | 12.3451 | 2/20/2020 10:23 |

The number of physical EVs was insufficient for evaluating the HHO solution on a larger scale. To address this limitation, we have extended the set by simulating 90 additional electric vehicles with the same features as the physical ones while maintaining the distribution of each car model (see Table 3).

*Table 3: The features of EVs used in the evaluation.*

| EV model | Battery capacity (kWh) | # EVs monitored | # EVs simulated |
|---|---|---|---|
| Renault ZOE | 22 | 3 | 35 |
| Renault ZOE | 41 | 4 | 45 |
| Nissan LEAF | 24 | 2 | 20 |

For the simulated EVs we generated the state of charge and the remaining number of charge/discharge cycles using different distributions. To model the SoC as a bounded variable between 20 and 80 percent we have used the beta distribution with shape parameters alpha=2 and beta=5 (see Figure 1). For the remaining number of charge/discharge cycles, we have used the Weibull Distribution with a shape parameter of 2 to represent wear-out failures and a scale of the distribution of 1000 representing the average remaining cycles.

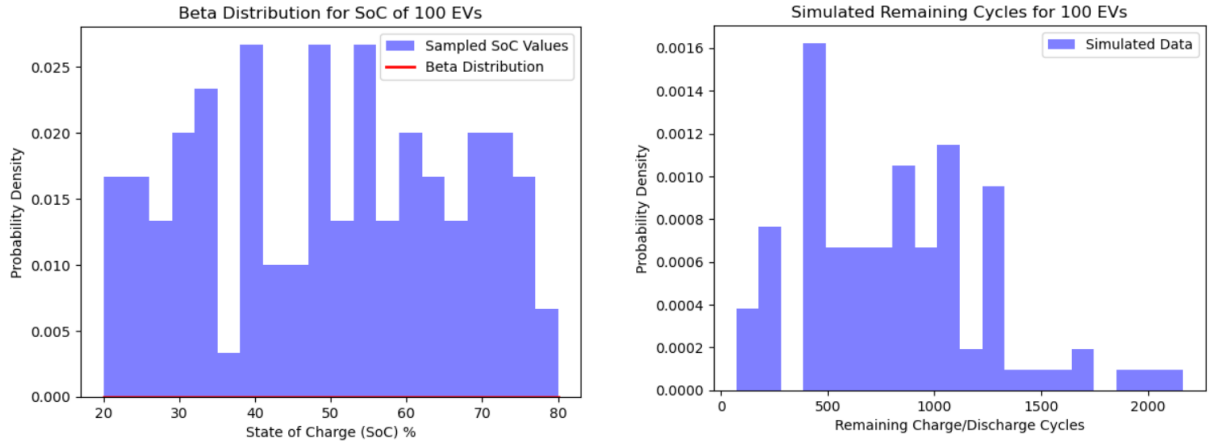

*Figure 1. SoC distribution on the left; remaining cycles of charge/discharge on the right*

To generate EV locations, we first defined a radius around each CS, then randomly generated latitude and longitude coordinates for each EV using a uniform distribution. Next, we checked that the generated locations are valid (i.e., the location of an EV does not overlap with the location of another EV or an obstacle), and if a location is invalid, it will be regenerated. We had access to 2 public CSs with 2 plugs located in Terni, each with the technical features described in Table 4.

*Table 4: Charging technical characteristics.*

| Model | SpotLink EVO |
|---|---|
| Nominal Voltage | 400 V AC THREE-PHASE |
| Nominal Frequency | 50 Hz |
| Current | 32 A |
| Sockets | two of Type 2 |
| Power Output | 22 kW |
| On-board Computer | Raspberry Pi 3 |

We have simulated 3 additional ones with the same technical features as the physical ones assigning virtual locations specified by latitude and longitude coordinates. In the process, we have leveraged on actual locations of public stations in Terni to which we did not have access via Emotion APIs. Figure 2 shows the map between CSs and EVs considered in our evaluation.

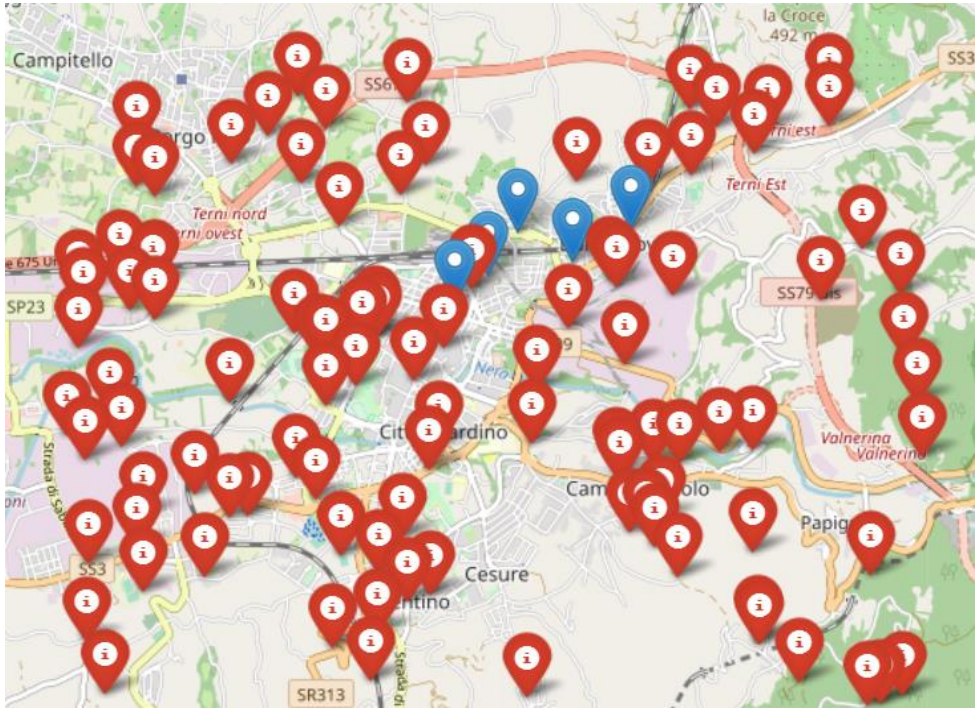

*Figure 2. The map of CSs (in blue) and EVs distribution (in red)*

To generate the drivers' preferences for the time interval when they want to charge or discharge vehicles, we have first defined the charging time slots. Next, we assigned the drivers' preferences for charging/discharging times to the time slots using a uniform distribution. Such an approach ensured an equal distribution of drivers' time preferences in all time slots. Finally, we adjust this initial distribution using specific strategies (i.e., more electric vehicles allocated for off-peak charging during the day or night due to low costs).

Our objective is to optimize EVs charging and discharging schedules to balance the local power grid. We have considered two specific scenarios: renewable energy integration and management of congestion (see Table 5).

*Table 5: Balancing services considered by our coordination model.*

| Service Name | | Time | Solution | HHO Means |
|---|---|---|---|---|
| Renewable integration | Renewable peak and not enough consumption | 8 Hours | Increase the demand to meet a forecasted renewable peak | Coordinated charge of EVs |
| Congestion Management | Consumption peak and not enough production | | Decrease the demand to avoid congestion and increase the generation | Coordinated discharge of EVs; Time delay of EVs charging |

To model the energy situation of each service we have generated for the microgrid an energy generation and a renewable energy production curve. For the production curve, we utilized PVWatts [52], a tool provided by the National Renewable Energy Laboratory, designed for creating energy production profiles for configurable photovoltaic systems. This allowed us to generate a realistic energy production curve based on the characteristics of the photovoltaic system specified in Table 6, considering the specifics of the microgrid's components and location, and variability in generation due and seasonal changes. For the consumption curve have used the actual consumption curve of households in Terni, Italy. By using these production and consumption curves, we were able to create scenarios that mimic congestion management challenges and renewable energy peak situations within the microgrid, enabling us to evaluate the effectiveness of EVs scheduling optimization strategies.

Table 6: Distributed PV systems features.

| Feature | Value |
|---:|:---|
| Generation Capacity | 100 kWh |
| Module Type | Monocrystalline (high efficiency of 20%) |
| Array Type | Fixed Tilt (open rack) |
| System Losses | 15% (losses due to shading, temperature, soiling, etc.) |
| Array Tilt Angle | 72° for summer months (taken from [53]) |
| Array Azimuth Angle | $42^0 + 15^0$ (Terni's latitude increased with 15 in the case of the summer energy production [54]) |
| Inverter Efficiency (%) | 96% (modern inverters with high efficiency) |
| Albedo | 0.2 (typical for ground surfaces) |
| Ground Coverage Ratio | 0.3 |
| Bifacial | No (assuming single-sided panels) |

## 5.1 Renewable integration

In this case, the EVs are coordinated like decentralized energy storage by charging batteries to store excess energy when the renewable generation exceeds the demand, helping to balance the grid. This stored energy can be used later when demand exceeds supply by coordinated discharge of the EVs batteries.

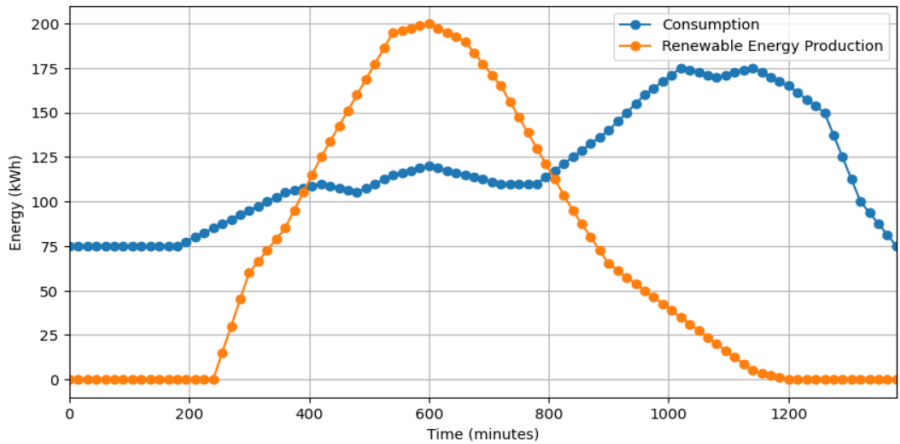

Figure 3. The peak of renewable energy exceeds the actual demand

Figure 3 shows the renewable generation and demand curves in this service. The solar panels on residential rooftops are producing electricity at their peak capacity being efficient due to the very good weather conditions. During midday, energy demand within the area is relatively low because many residents are at work or school, and some industrial activities are reduced.

HHO solution will coordinate the charging of EVs connected to the microgrid, to increase energy demand between 7 AM and 2 PM. We have set the population size to 20 individuals, the maximum number of iterations to 100, and an archive dimension to 5. These values are the most effective combination for our EV scheduling problem. They were determined by a trial-and-error procedure where we systematically varied the population size, the number of iterations, and the archive dimension and analyzed their impact on the algorithm performance.

The results are presented in Figure 4. As can be seen, the determined charging schedule manages to increase the energy demand to successfully store most of the renewable energy surplus from the microgrid. Therefore, the fleet of EVs is effectively managed as a decentralized energy storage to align with renewable energy peak maximizing the utilization of locally produced energy. This not only enhances the reliability of the microgrid-level power supply but also contributes to reducing the carbon footprint as we have shown below.

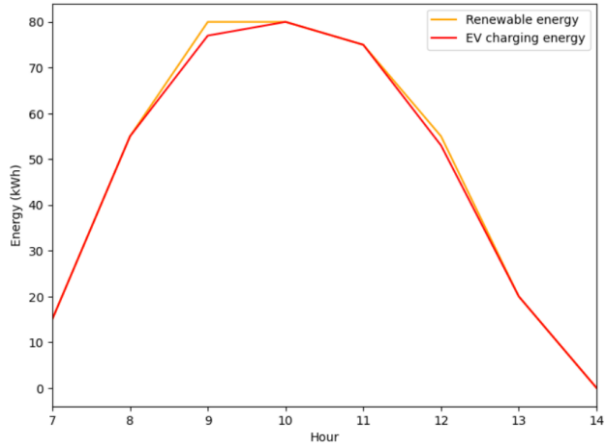

*Figure 4. Renewable energy and additional demand of EVs charging.*

The Pearson coefficient was used as a quantitative measure to assess how well the scheduling of each EV's charging aligns with the microgrid's energy production. A coefficient value closer to 1 means that the scheduling is effective in utilizing available renewable energy surplus, while if the coefficient is close to 0, it suggests a mismatch between energy production and total EV consumption. We compared the renewable energy surplus in the microgrid with the additional energy consumption due to the HHO schedule of EV charging. The coefficient value obtained is 0.99 very close to 1 reflecting the performance of HHO in generating EV schedules to follow the renewable.

Figure 5 illustrates the energy in the hourly distribution of EVs charged according to the schedule generated by the extended HHO-based optimization algorithm. As can be seen, it successfully manages to fill the hourly energy demand while considering each EV's technical requirements for charging and its actual state of charge. The total energy charged by EVs, $e_{storage}(T)$ is about 375 KWh.

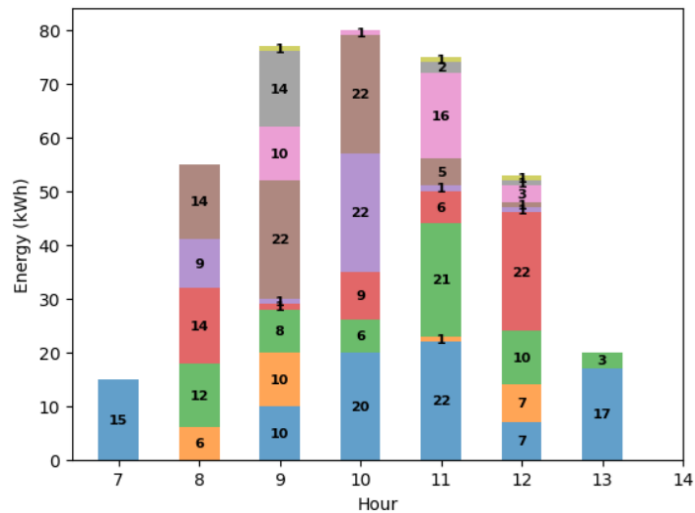

*Figure 5. The distribution of EVs per hour of the service and amount of energy charged.*

For the EVs distribution, 41 vehicles out of the 100 vehicles have been scheduled for charging. Table 7 presents how the time and distance preferences are met in scheduling. The results show that the HHO-based optimization algorithm has effectively considered the drivers' charging time preferences maintaining an acceptable deviation of a maximum of 2 hours from the preferred time interval in most of the cases. Only 1 EV has been scheduled at a time more than 2 hours earlier or later than the preferred time interval. Regarding the distance constraints, it can be noticed that for 38 EVs the constraints have been fully met, while only for 3 EVs the distance constraint has been slightly relaxed.

*Table 7: Consideration of users' time preferences concerning EVs schedule.*

| Preference type | Preferences fulfillment level | #EVs |
|---|---|---|
| Time | Fully met | 36 |
| | 1-hour deviation | 4 |
| | 2-hours deviation | 0 |
| | More than 2-hours deviation | 1 |
| Distance | Fully met | 38 |
| | Minimum deviation | 3 |

Finally, we have determined the impact on carbon footprint reduction due to the integration of renewable energy we have by referring to the energy mix. As a result of increasing the local usage of renewable energy, the microgrid avoids the import of electricity from the main grid during other times of the day when demand exceeds generation. For estimation, we used the energy mix of Italy [55] ($MIX_{gCo2/kWh} = 363\ g\ CO2eq/kWh$), in which even if the use of fossil gas in energy production still predominates, the use of renewable energy sources has registered a substantial increase:

$$\Delta_{CO2}(T) = MIX_{gCo2/kWh} * e_{storage}(T) \quad (52)$$

This resulted in a total of around 136kg of CO2 saved at the microgrid level.

## 5.2. Congestion management

In this case, the EVs are coordinated to deploy additional decentralized generation capacity to prevent grid congestion. Figure 6 shows a zoom on Figure 3, focusing on the period between 2 PM and 11 PM when the energy imported by the microgrid together with the renewable energy generated does not meet the required energy demand. It can be noticed that after 2 PM the energy demand increases with a peak between 4 PM - 7 PM which correlates with people's behaviour at home using intensively electric devices for their household activities.

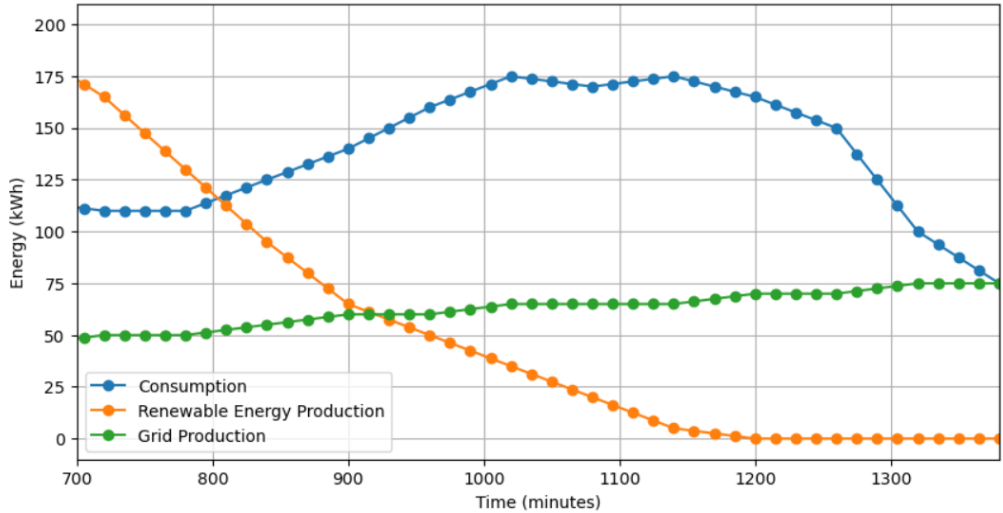

*Figure 6. Peak of energy consumption exceeding the energy production*

The HHO solution will coordinate the discharging of EVs in the microgrid to increase the quantity of available energy to meet the energy demand between 3 PM and 10 PM and avoid congestion. Like in the renewable integration scenario, we have applied a trial-and-error procedure to determine the most effective values of the algorithm's adjustable parameters and as a result, we have set the population size to 20 individuals, the maximum number of iterations to 100, and an archive dimension to 5. The results presented in Figure 7 show that the determined charging schedule successfully increased the available energy to meet the energy demand.

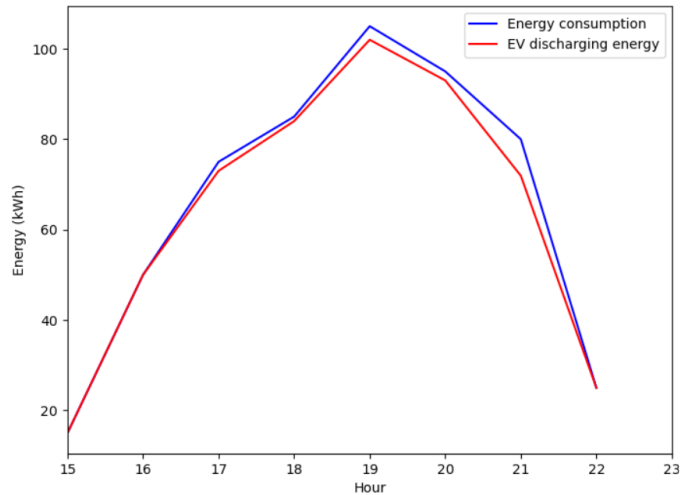

Figure 7. EVs discharged energy to support energy demand.

In the congestion management scenario, we have used the Pearson coefficient as a quantitative measure to assess how well the scheduling of each EV's discharging aligns with the microgrid's demand energy. We have obtained a coefficient value equal to 0.99, very close to 1 reflecting the performance of the HHO optimization algorithm in coordinating the EVs' discharge.

Figure 8 illustrates the energy discharged by the EVs in each hour according to the schedule generated by the extended HHO-based optimization algorithm. As can be seen, it successfully manages to fill the hourly energy demand while considering each EV's technical requirements for discharging and its actual state of charge. The total energy discharged be EVs, $e_{discharge}(T)$ is about 515 kWh.

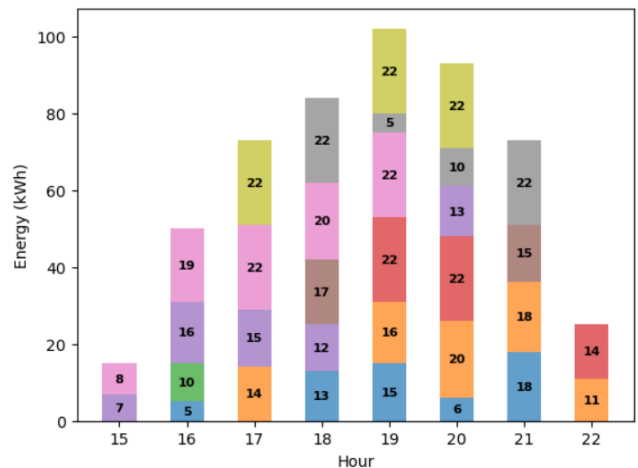

Figure 8. The distribution of EVs per hour of the service and amount of energy discharged.

For EVs schedule depicted in Figure 8, 33 vehicles out of the 41 vehicles previously charged in the renewable integration scenario have been scheduled for discharge during the interval of congestion management. Table 7 shows that the time and distance preferences are effectively met in scheduling. HHO-based optimization algorithm has considered the drivers' discharging time preferences maintaining an acceptable deviation of a maximum of 2 hours from the preferred time interval in most of the cases. Only 5 EV have been scheduled at a time more than 2 hours earlier or later than the preferred time interval. Regarding the distance constraints, it can be noticed that 30 EVs have been scheduled at a CS in the maximum range preferred by each driver, and only 3 EVs have been scheduled at a CS located at a distance with a minimum deviation from the maximum preferred one.

Table 8: Consideration of users' time and distance preferences in relation to EVs schedule.

| Preference type | Preferences fulfilment level | #EVs |
|---|---|---|
| Time | Fully met | 15 |
| | 1-hour deviation | 11 |
| | 2-hours deviation | 2 |
| | More than 2-hours deviation | 5 |
| Distance | Fully met | 30 |
| | Minimum deviation | 3 |

Finally, we have determined the energy flexibility gain of our solution due to the EVs coordination. By providing additional energy to the grid through discharging the EVs, the quantity of energy available in the grid increases and can satisfy the energy demand that cannot be satisfied by the grid solely. For estimating the energy flexibility, we have used the formula below [56]:

$$EF(\%) = \frac{EnergyEV - EnergyBaseline}{EnergyPeak} = \frac{300.21}{1043.92} * 100 = 28.75\% \tag{53}$$

where baseline energy is the quantity of energy that would have been discharged by EVs in the absence of our solution, while *energy peak* represents the maximum electricity value available in the microgrid. It can be noticed that in our case we obtain a significant increase in energy flexibility of approximately 28 %.

## 6. Discussion

To evaluate the performance of our coordination solution based on HHO algorithm, we performed an analysis covering several aspects: (i) convergence, (ii) spread and distribution of solutions from the approximate Pareto front, and (iii) the quality of solutions.

The coordination algorithm's convergence was analysed to understand the impact of the number of iterations over the algorithm progress and to what extent it reaches satisfactory solutions for our EVs scheduling problem. We have determined the distance between the Pareto front and its representation in the objective space, and the spread of the approximated front [57]. This is especially important in multi-objective optimization, as it involves finding solutions that trade-off among conflicting objectives. We have used the Generational Distance (GD) [57] to measure how close the solutions obtained by our algorithm (i.e., the approximate Pareto front) are to the ideal Pareto front. The Generational Distance considers the average quality of the approximated front solutions. We have also used the Maximum Pareto Front Error [57] to measure the worst-case error between the approximate Pareto front determined by our HHO algorithm and the ideal Pareto front. The metric determines the distance between the solutions found by our algorithm and the ideal Pareto front thus lower values show better convergence performance. The values we have obtained for the Generational Distance and Maximum Pareto Front Error metrics while varying the number of algorithm iterations are illustrated in Table 9. It can be noticed that by increasing the number of iterations, the values of the Generational Distance and Maximum Pareto Front Error metric decrease. For example, when 100 iterations are used, the Generational Distance metric has a value close to 0 indicating that the solutions obtained by our algorithm are close to the ideal Pareto front. Finally, we evaluated the quality of the solutions obtained in the approximate Pareto front. We have used the Hyper-Volume [58] to determine the coverage of the solutions in the objective space. For the Hyper-Volume computation, we have used as reference point the nadir point. We have obtained a value of 0.99 indicating the good algorithm's convergence for our EVs scheduling problem.

*Table 9: Converge, Generational Distance and Maximum Pareto Front Error.*

| Number of iterations | Generational Distance | Maximum Pareto Front Error |
|---|---|---|
| 70 | 3.83 | 6.13 |
| 80 | 2.5 | 6 |
| 90 | 2.43 | 4.24 |
| 100 | 0.35 | 1.01 |

The second aspect analysed is the distribution and diversity of solutions within a population that reflect the balance between the exploration and exploitation phases of the algorithm. The Spacing metric quantifies how well-distributed the non-dominated solutions are in the search space. Table 10 illustrates the variation of spacing when the number of iterations increases. For 100 iterations, we computed a low value (i.e., 0.77) for the spacing measure, indicating a uniform distribution of non-dominated solutions. To determine the percentage of non-dominated solutions in the population, we calculated the Ratio of Non-Dominated Individuals as referenced in [57]. In our case, we obtained a low average ratio of 5%, influenced by the simultaneous minimization of many objectives (i.e., 8 objective functions) and the satisfaction of multiple constraints.

*Table 10: Solutions spacing variation with the algorithm iterations.*

| Number of iterations | Spacing |
|---|---|
| 70 | 2.87 |
| 80 | 2.65 |
| 90 | 1.73 |
| 100 | 0.77 |

The final aspect we examined is the quality of the solutions returned by the algorithm. We have used the Parallel Coordinates Plot to determine and evaluate the degree to which our algorithm manages to balance the trade-offs between different objectives. In this case, each polyline represents a solution that spans all parallel axes, with each axis intersection representing the corresponding objective value. Figures 9 and 10 illustrate the Parallel Coordination Plots for the non-dominant solutions present in the archive at the end of 100 iterations (we chose to plot these solutions because, for 100 iterations, the proposed algorithm provides the best values for the Generational Distance, Maximum Pareto Front Error, and Spacing metrics) and for the solutions in the Pareto front. The results show that the proposed algorithm has a good convergence, and the fitness function values for the non-dominated archive solutions have almost the same range as in the case of the Pareto front solutions, with acceptable deviations.

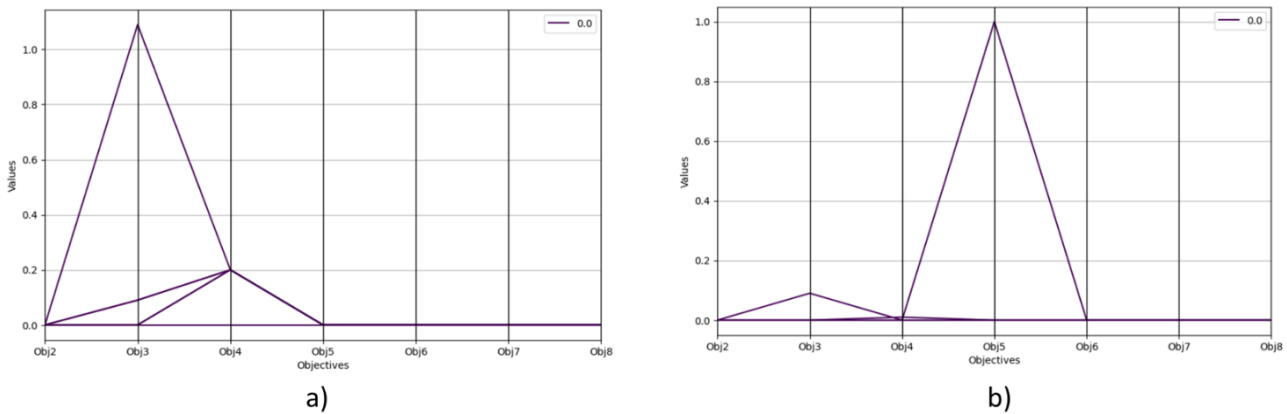

*Figure 9. Parallel Coordination Plot for: a) the non-dominated archive solutions; b) Pareto front solutions*

To determine visually the performance of the obtained solutions against each objective we have used a Heat map. Figure 10 illustrates the heatmap generated for our solution, for the non-dominated solutions part of the archive. In the representation, the X-axis represents the objective functions, and the Y-axis represents the values for the objective functions for each solution in the approximate Pareto front. Darker areas in certain columns indicate solutions that perform poorly for certain objectives. Lighter areas in certain columns denote solutions that work well for specific objectives. By comparing colours row by row, we can analyse trade-offs between different objectives. If the improvement of one objective (resulting in a darker cell) leads to the deterioration of another objective (resulting in a lighter cell), as we our case, we can conclude that there is a trade-off between these objectives and, consequently, better solutions are obtained.

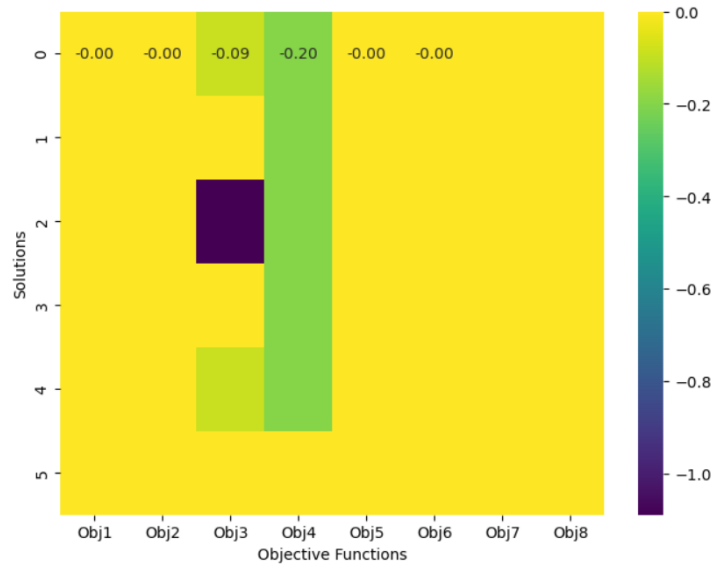

Figure 10. Heatmap of solutions and objective functions values.

## 7. Conclusion

In this paper, we have proposed a coordination model for electric vehicles (EVs) that optimizes the day-ahead energy usage of EVs to provide balancing services in the local grid. This is achieved by storing surplus energy in EVs and discharging it during periods of energy deficit enabling the fleet of EVs to act as a decentralized energy storage. We defined the day-ahead coordination of EVs as a multi-criteria optimization problem, where each time instance represents an objective of aligning energy demand and supply while considering the local operational constraints of EVs and CSs, user time-of-use constraints, and the constraints of the balancing service. We use the Harris hawk's optimization (HHO) algorithm, by having the EVs schedules represented as energy matrices indicating charging and discharging activities at specific stations during different periods. The algorithm's iterations adjust the EVs schedules associated with the position of individuals from the population through exploration and exploitation operations to also meet the constraints.

The results obtained in the context of an e-mobility service are promising showing that our proposed solution can coordinate the energy charging and discharging operations of a fleet of EVs to balance the energy in the microgrid while adhering with minimum deviations to the technical operational and time of use constraints. Consequently, the EVs can use as much as possible the renewable energy surplus during the daily peak and use their flexibility by discharging the stored energy during the moments of energy deficit supporting the grid operation. Moreover, the evaluation results show our solution's quality and efficacy as well as good convergence.

In future work, we plan to integrate model-free machine learning solutions to enhance the adaptability capabilities of the coordination model to the grid chaining conditions as well as to the user behavior and preferences. In this way, we can improve the model's ability to learn and respond to dynamic changes in the microgrid and user behavior enabling real-time adaptation and optimization. Also, work needs to be done at the fronter of social sciences with energy and computer science fields to explore new ways to incentivize EV users to participate in the coordination scheme by incorporating their feedback and designing effective incentive structures that can go beyond the purely financial owns towards community building.

## Acknowledgment

This work has been conducted within the BRIGHT project grant number 957816 funded by the European Commission as part of the H2020 Framework Programme and the DEDALUS project grant number 101103998 funded by the European Commission as part of the Horizon Europe Framework Programme.

## References

[1] Trends in electric light-duty vehicles, Global EV Outlook 2023, Available online: https://www.iea.org/reports/global-ev-outlook-2023/trends-in-electric-light-duty-vehicles [Accessed on 22.11.2023]

[2] Alanazi, F. Electric Vehicles: Benefits, Challenges, and Potential Solutions for Widespread Adaptation. Appl. Sci. 2023, 13, 6016.

[3] Electric vehicles: Setting a course for 2030, Deloitte report, 2020, Available online: https://www2.deloitte.com/us/en/insights/focus/future-of-mobility/electric-vehicle-trends-2030.html [Accessed on 22.11.2023]

[4] Trends in batteries, Global EV Outlook, 2023, Available online: https://www.iea.org/reports/global-ev-outlook-2023/trends-in-batteries [Accessed on 22.11.2023]

[5] A. Qazi et al., "Towards Sustainable Energy: A Systematic Review of Renewable Energy Sources, Technologies, and Public Opinions," in IEEE Access, vol. 7, pp. 63837-63851, 2019

[6] Regulation (EU) 2023/851 of the European parliament and of the council of 19 April 2023, Available online: https://eur-lex.europa.eu/legal-content/EN/TXT/?uri=CELEX:32023R0851 [Accessed on 22.11.2023]

[7] California's Zero Emissions Vehicle Rule and Its Nationwide Impacts, 2022, Available online: https://www.americanactionforum.org/insight/californias-zero-emissions-vehicle-rule-and-its-nationwide-impacts/ [Accessed on 22.11.2023]

[8] Khalifa, A.A.; Ibrahim, A.-J.; Amhamed, A.I.; El-Naas, M.H. Accelerating the Transition to a Circular Economy for Net-Zero Emissions by 2050: A Systematic Review. Sustainability 2022, 14, 11656.

[9] M. Browning, J. McFarland, J. Bistline, G. Boyd, M. Muratori, M. Binsted, C. Harris, T. Mai, G. Blanford, J. Edmonds, A. A. Fawcett, O. Kaplan, J. Weyant, Net-zero $CO_2$ by 2050 scenarios for the United States in the Energy Modeling Forum 37 study, Energy and Climate Change, Volume 4, 2023

[10] F. Palmiotto, Y. Zhou, G. Forte, M. Dicorato, M. Trovato, L.M. Cipcigan, A coordinated optimal programming scheme for an electric vehicle fleet in the residential sector, Sustainable Energy, Grids and Networks, Volume 28, 2021

[11] M. S. Mastoi, S. Zhuang, H. Mudassir Munir, M. Haris, M. Hassan, M. Alqarni, B. Alamri, A study of charging-dispatch strategies and vehicle-to-grid technologies for electric vehicles in distribution networks, Energy Reports, Volume 9, 2023

[12] S. Ray, K. Kasturi, S. Patnaik, M. R. Nayak, Review of electric vehicles integration impacts in distribution networks: Placement, charging/discharging strategies, objectives and optimisation models, Journal of Energy Storage, Volume 72, Part D, 2023

[13] C. Gschwendtner, S. R. Sinsel, A. Stephan, Vehicle-to-X (V2X) implementation: An overview of predominate trial configurations and technical, social and regulatory challenges, Renewable and Sustainable Energy Reviews, Volume 145, 2021

[14] X. Tian, B. Cheng, H. Liu, V2G optimized power control strategy based on time-of-use electricity price and comprehensive load cost, Energy Reports, Volume 10, 2023

[15] S. Chen, G. Sun, Z. Wei, D. Wang, Dynamic pricing in electricity and natural gas distribution networks: An EPEC model, Energy, Volume 207, 2020, 118138

[16]  P. Barman, L. Dutta, S. Bordoloi, A. Kalita, P. Buragohain, S. Bharali, B. Azzopardi, Renewable energy integration with electric vehicle technology: A review of the existing smart charging approaches, Renewable and Sustainable Energy Reviews, Volume 183, 2023

[17]  El-Bayeh, C.Z.; Alzaareer, K.; Aldaoudeyeh, A.-M.I.; Brahmi, B.; Zellagui, M. Charging and Discharging Strategies of Electric Vehicles: A Survey. World Electr. Veh. J. 2021, 12, 11.

[18]  T. U. Solanke, V. K. Ramachandaramurthy, J. Y. Yong, J. Pasupuleti, P. Kasinathan, A. Rajagopalan, A review of strategic charging–discharging control of grid-connected electric vehicles, Journal of Energy Storage, Volume 28, 2020

[19]  Ke, S., Chen, L., Yang, J., Li, G., Wu, F., Ye, L., Wei, W., Wang, Y.: Vehicle to everything in the power grid (V2eG): A review on the participation of electric vehicles in power grid economic dispatch. Energy Convers. Econ. 3, 259–286 (2022).

[20]  Pritima, D., Rani, S.S., Rajalakshmy, P., Kumar, K.V., Krishnamoorthy, S. (2022). Artificial Intelligence-Based Energy Management and Real-Time Optimization in Electric and Hybrid Electric Vehicles. In: Kathiresh, M., Kanagachidambaresan, G.R., Williamson, S.S. (eds) E-Mobility. EAI/Springer Innovations in Communication and Computing. Springer, Cham.

[21]  K. Sevdari, L. Calearo, P. B. Andersen, M. Marinelli, Ancillary services and electric vehicles: An overview from charging clusters and chargers technology perspectives, Renewable and Sustainable Energy Reviews, Volume 167, 2022, 112666

[22]  Y. Zhou and X. Li, "Vehicle to grid technology: A review," 2015 34th Chinese Control Conference (CCC), Hangzhou, China, 2015, pp. 9031-9036

[23]  Y. Ding, X. Li, S. Jian, Modeling the impact of vehicle-to-grid discharge technology on transport and power systems, Transportation Research Part D: Transport and Environment, Volume 105, 2022, 103220

[24]  Z. Wan, H. Li, H. He and D. Prokhorov, "Model-Free Real-Time EV Charging Scheduling Based on Deep Reinforcement Learning," in IEEE Transactions on Smart Grid, vol. 10, no. 5, pp. 5246-5257, Sept. 2019

[25]  A. K. Kalakanti and S. Rao. 2023. Computational Challenges and Approaches for Electric Vehicles. ACM Comput. Surv. 55, 14s, Article 311 (December 2023)

[26]  J. García-Álvarez, M. A. González, C. R. Vela, Metaheuristics for solving a real-world electric vehicle charging scheduling problem, Applied Soft Computing, Volume 65, 2018

[27]  A. A. Heidari, S. Mirjalili, H. Faris, I. Aljarah, M. Mafarja, H. Chen, Harris hawks optimization: Algorithm and applications, Future Generation Computer Systems, Volume 97, 2019, Pages 849-872, ISSN 0167-739X

[28]  H. Song, C. Liu, M. Jalili, X. Yu, P.r McTaggart, Multi-objective Scheduling of Electric Vehicle Charging/Discharging with Time of Use Tariff, IEEE Transactions on Smart Grid, 2021.

[29]  Morais, H.; Sousa, T.; Castro, R.; Vale, Z. Multi-Objective Electric Vehicles Scheduling Using Elitist Non-Dominated Sorting Genetic Algorithm. Appl. Sci. 2020, 10, 7978.

[30]  Jin, H.; Lee, S.; Nengroo, S.H.; Har, D. Development of Charging/Discharging Scheduling Algorithm for Economical and Energy-Efficient Operation of Multi-EV Charging Station. Appl. Sci. 2022, 12, 4786.

[31]  Poniris, S.; Dounis, A.I. Electric Vehicle Charging Schedules in Workplace Parking Lots Based on Evolutionary Optimization Algorithm. Energies 2023, 16, 221.

[32]  Piamvilai, N.; Sirisumrannukul, S. Optimal Scheduling of Movable Electric Vehicle Loads Using Generation of Charging Event Matrices, Queuing Management, and Genetic Algorithm. Energies 2022, 15, 3827.

[33]  N. T. Milas, D. A. Mourtzis, P. I. Giotakos and E. C. Tatakis, "Two-Layer Genetic Algorithm for the Charge Scheduling of Electric Vehicles," 2020 22nd European Conference on Power Electronics and Applications, Lyon, France, 2020, pp. P.1-P.10, doi: 10.23919/EPE20ECCEEurope43536.2020.9215685.

[34]  R. Wang, J. Mu, Z. Sun, J. Wang, A. Hu, NSGA-II multi-objective optimization regional electricity price model for electric vehicle charging based on travel law, Energy Reports, Volume 7, Supplement 7, 2021, Pages 1495-1503, ISSN 2352-4847.

[35] S. Abdullah-Al-Nahid, T. Ahmed Khan, A. Taseen, T. Jamal, T. Aziz, A novel consumer-friendly electric vehicle charging scheme with vehicle to grid provision supported by genetic algorithm based optimization, Journal of Energy Storage, Volume 50, 2022, 104655, ISSN 2352-152X
[36] Yang, J.; He, L.; Fu, S. An improved PSO-based charging strategy of electric vehicles in electrical distribution grid. Appl. Energy 2014, 128, 82–92.
[37] Konstantinidis, G.; Kanellos, F.D.; Kalaitzakis, K. A Simple Multi-Parameter Method for Efficient Charging Scheduling of Electric Vehicles. Appl. Syst. Innov. 2021, 4, 58.
[38] Fernandez, G.S.; Krishnasamy, V.; Kuppusamy, S.; Ali, J.S.; Ali, Z.M.; El-Shahat, A.; Abdel Aleem, S.H.E. Optimal Dynamic Scheduling of Electric Vehicles in a Parking Lot Using Particle Swarm Optimization and Shuffled Frog Leaping Algorithm. Energies 2020, 13, 6384. https://doi.org/10.3390/en13236384
[39] Mohammad, A.; Zuhaib, M.; Ashraf, I.; Alsultan, M.; Ahmad, S.; Sarwar, A.; Abdollahian, M. Integration of Electric Vehicles and Energy Storage System in Home Energy Management System with Home to Grid Capability. Energies 2021, 14, 8557. https://doi.org/10.3390/en14248557
[40] Fang B, Li B, Li X, Jia Y, Xu W and Liao Y (2021) Multi-Objective Comprehensive Charging/Discharging Scheduling Strategy for Electric Vehicles Based on the Improved Particle Swarm Optimization Algorithm. Front. Energy Res. 9:811964. doi: 10.3389/fenrg.2021.811964
[41] N. Wang, B. Li, Y. Duan, S. Jia, A multi-energy scheduling strategy for orderly charging and discharging of electric vehicles based on multi-objective particle swarm optimization, Sustainable Energy Technologies and Assessments, Volume 44, 2021, 101037, ISSN 2213-1388
[42] Savari, G.F.; Krishnasamy, V.; Sugavanam, V.; Vakesan, K. Optimal charging scheduling of electric vehicles in micro grids using priority algorithms and particle swarm optimization. Mob. Netw. Appl. 2019, 24, 1835–1847.
[43] Sowmya R, V. Sankaranarayanan, Optimal vehicle-to-grid and grid-to-vehicle scheduling strategy with uncertainty management using improved marine predator algorithm, Computers and Electrical Engineering, Volume 100, 2022, 107949, ISSN 0045-7906.
[44] García Álvarez, J.; González, M.Á.; Rodríguez Vela, C.; Varela, R. Electric Vehicle Charging Scheduling by an Enhanced Artificial Bee Colony Algorithm. Energies 2018, 11, 2752. https://doi.org/10.3390/en11102752
[45] Hernandez-Arauzo, A.; Puente, J.; Varela, R.; Sedano, J. Electric Vehicle Charging under Power and Balance Constraints as Dynamic Scheduling. Comput. Ind. Eng. 2015, 85, 306–315.
[46] Ahmadi, M., Hosseini, S.H. & Farsadi, M. Optimal Allocation of Electric Vehicles Parking Lots and Optimal Charging and Discharging Scheduling using Hybrid Metaheuristic Algorithms. J. Electr. Eng. Technol. 16, 759–770 (2021). https://doi.org/10.1007/s42835-020-00634-z
[47] Abdel-Hakim, A.E.; Abo-Elyousr, F.K. Heuristic Greedy Scheduling of Electric Vehicles in Vehicle-to-Grid Microgrid Owned Aggregators. Sensors 2022, 22, 2408. https://doi.org/10.3390/s22062408
[48] Limmer, S.; Varga, J.; Raidl, G.R. Large Neighborhood Search for Electric Vehicle Fleet Scheduling. Energies 2023, 16, 4576. https://doi.org/10.3390/en16124576
[49] Alinejad, M.; Rezaei, O.; Habibifar, R.; Azimian, M. A Charge/Discharge Plan for Electric Vehicles in an Intelligent Parking Lot Considering Destructive Random Decisions, and V2G and V2V Energy Transfer Modes. Sustainability 2022, 14, 12816. https://doi.org/10.3390/su141912816
[50] A. Lipowski, D. Lipowska, Roulette-wheel selection via stochastic acceptance, Physica A: Statistical Mechanics and its Applications, Elsevier, 2012, vol. 391(6), pages 2193-2196.
[51] EMOTION company, Available online: https://emotion-team.com/ [Accessed on 24.11.2023]
[52] PVWatts Calculator, Available online: https://pvwatts.nrel.gov/ [Accessed on 24.11.2023]
[53] A. R. Salih, Seasonal Optimum Tilt Angle of Solar Panels for 100 Cities in the World, March 2023Al-Mustansiriyah Journal of Science 34(1):104-110, DOI:10.23851/mjs.v34i1.1250
[54] Tilt & Azimuth Angle: Finding the Optimal Angle to Mount Your Solar Panels, Unbound Solar, 2018, Available online: https://unboundsolar.com/blog/solar-panel-azimuth-angle [Accessed on 24.11.2023]

[55] Energy mix of Italy, 2023, Available online: https://www.nowtricity.com/country/italy/#:~:text=Quick%20stats%20about%20Italy,Coal%20usage%20was%2011.7%25 [Accessed on 24.11.2023]

[56] T. Cioara, I. Anghel, M. Bertoncini, I. Salomie, D. Arnone, M. Mammina, T. Velivassaki, M. Antal, Optimized Flexibility Management enacting Data Centres Participation în Smart Demand Response Programs, Future Generation Computer Systems, Volume 78, Part 1, January 2018, Pages 330-342

[57] C Audet, J Bigeon, D Cartier, S Le Digabel, L Salomon, Performance indicators in multiobjective optimization, European journal of operational research 292 (2), 397-422, 2021.

[58] A. P. Guerreiro, C. M. Fonseca, and L. Paquete. The Hypervolume Indicator: Computational Problems and Algorithms. ACM Comput. Surv. 54, 6, July 2022.